# Iris Recognition: Inherent Binomial Degrees of Freedom

J. Michael Rozmus
Principal Algorithm Engineer, Eyelock LLC
mrozmus@eyelock.com
IEEE Life Senior Member

**Abstract**

*The distinctiveness of the human iris has been measured by first extracting a set of features from the iris – an encoding – and then comparing these encoded feature sets to determine how distinct they are from one another. For example, John Daugman measures the distinctiveness of the human iris at 244 degrees of freedom, that is, Daugman's encoding maps irises into the equivalent of $2^{244}$ distinct possibilities [2]. This paper shows by direct pixel-by-pixel comparison of high-quality iris images that the inherent number of degrees of freedom embodied in the human iris – independent of any encoding – is at least 536. When the resolution of these images is gradually reduced, the number of degrees of freedom decreases smoothly to 123 for the lowest resolution images tested.*

## 1. Introduction

In 1993 [1], John Daugman described his very successful method of iris biometric identification by noting the presence or absence of iris features extracted by spatial filtering of normalized iris images with wavelet kernels. The proportion of non-matching features is taken to be a Hamming distance. An imposter histogram of the Hamming distances between different eyes has a mean of 0.5 and a relatively small standard deviation. Daugman shows that this histogram closely follows the discrete binomial probability density function having the same mean and standard deviation (see figure 6 in [1]).

The discrete binomial probability density function is given by the following equation:

$$f(x) = \frac{N!}{k!(N-k)!} \, p^k (1-p)^{(N-k)} \quad (1)$$

where N is the number of Bernoulli trials, p is the probability of success for a single trial, k is the number of successes in the N trials, and x = k/N. The standard deviation as a proportion of N is $\sigma = (p(1-p)/N)^{½}$ and thus

$$N = \frac{p\,(1-p)}{\sigma^2} \quad (2)$$

Daugman referred to N as the number of independent "degrees of freedom" (see page 1155 in [1]). In 2007 [2], he calculated the degrees of freedom for an imposter distribution of 200 billion iris comparisons to be 244. The value of the binomial degrees of freedom is a measure of distinctiveness of one iris from all others.

The complex structure of the human iris is a signal that communicates the presence of a specific person – biometric identification. John Daugman's method of encoding and decoding that signal, and all of the related methods that dominate commercial iris recognition today, extract phase information from that signal – detecting the relative position of peaks, valleys, and slopes in the iris texture. Daugman explains this approach as follows (see page 23 in [3]):

"Only phase information is used for recognizing irises because amplitude information is not very discriminating, and it depends upon extraneous factors such as imaging contrast, illumination, and camera gain."

In other words, current iris encoding methods discard amplitude information because of the practical difficulties of reliably extracting that information. But consider a comparison to the field of communications engineering. Proven methods of communicating a signal in an electronic channel include both amplitude modulation and phase shifting. Both amplitude and phase can carry significant information.

This paper reports an investigation of the degrees of freedom inherent in the complex structure of the human iris. Avoiding any encoding or feature extraction that reduces information content, irises are compared pixel-by-pixel, retaining both amplitude and phase information.

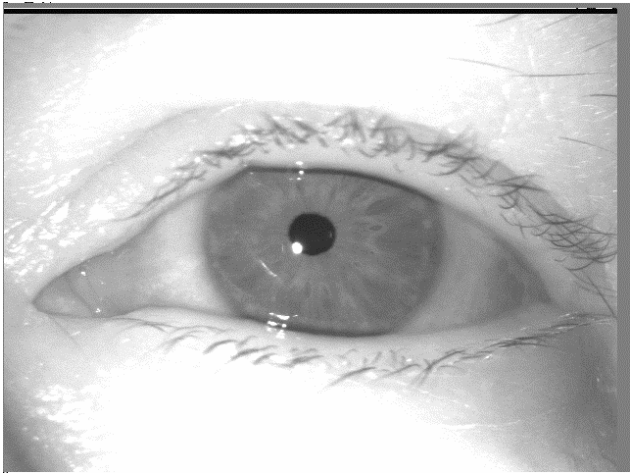

Figure 1: High-quality iris image sample (courtesy of Notre Dame University, See reference [4])

## 2. Overview of the Investigation

The investigation begins with 3000 very high-quality iris images courtesy of Notre Dame University [4] (see figure 1). A proportion of these were selected for brightness (good signal to random noise ratio) and maximum pupil radius (sufficient radial resolution). All of the 1382 selected images are normalized to a 128x960 rectangular format with the same median intensity and then masked to cover portions of the iris obscured by eyelid/eyelash occlusion and bright specular reflections (see figure 2).

The resulting normalized and masked iris images are compared to one another pixel-by-pixel for the portion of the iris visible in both. The Hamming distance is the proportion of that common visible iris area that does not have matching values of pixel intensity (within a margin of error). The result is an imposter histogram of 1.9 million iris comparisons. This histogram is matched to its corresponding binomial distribution to measure the degrees of freedom.

The iris images are further processed to obtain a graduated sequence of reduced resolution image sets. For each set, the process of the preceding paragraph is repeated to obtain a graph of degrees of freedom versus resolution.

## 3. High-Quality Iris Images

The images used in this investigation have the following characteristics [4]:
- 3000 unique 8-bit 480x640 eye images from 1500 individuals captured with an LG 4000 iris camera
- High resolution: iris diameter = 198 to 288 pixels (exceeding latest ISO standards for high quality iris images – page 19 in [5])
- Apparently excellent sharpness and signal to noise ratio
- A large range of pupil size

From these 3000 images, 1382 images were selected for a median iris pixel intensity >= 70 (good signal to random noise ratio) and a maximum pupil radius of 52 pixels (sufficient radial resolution on the iris).

## 4. Normalized Iris Format

The iris intensity values in the selected images were mapped to polar coordinates in the manner of Daugman's rubber sheet model (see page 25 in [3]). In order to preserve all of the resolution of the original images, there are 128 pixels in the radial direction and 960 pixels in the angular direction.

Consider the example of Figure 2. The top edge is the pupil-iris boundary. The bottom edge is the iris-sclera boundary. The vertical position is the radial location in the original image (Figure 1). The horizontal location is the angular position in the original image. The left edge and the right edge meet at the 3 o'clock position in the original view of the iris. The black regions mask the occluding eyelids, eyelashes, and specular reflections.

## 5. Algorithms

The algorithms described below are implemented in Matlab with the Image Processing Toolbox (version 2020a). The conversion from 480x640 eye images to the 128x960 rectangular normalized iris format and the calculation of the occlusion mask were done with proprietary software developed by Eyelock LLC and these algorithms are not described below.

### 5.1. Intensity Normalization

To correct for variation in illumination and the mean reflectance of irises, the intensity of iris pixels in the 128x960 format are normalized to a median value of 127 (approximately half of the maximum value of 255) by multiplying each unmasked pixel by (127 / (median value of all unmasked pixels). The variance of the intensity of iris pixels was not normalized, preserving the range of iris signal amplitudes in the 1382 selected irises.

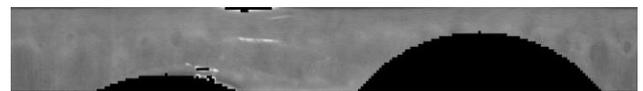

Figure 2: 128x960 Normalized iris with mask made from image of Figure 1

## 5.2. Comparing Irises

The comparison calculations were done with double floating point values. Each pixel location that was unmasked in both normalized and masked iris images was checked to see if the intensity values matched. A match occurred when the absolute value of the difference of the two pixel values was less than 0.5 / 255.

The Hamming distance is defined to be the proportion of iris pixels, unmasked in both images, that do not match.

## 5.3. Resolution Reduction

Matlab function *imresize* with its default bicubic interpolation algorithm was used to make lower resolution versions of the 1382 normalized and masked irises. The images were scaled down in resolution in both dimensions by 80, 50, 40, 30, 20, 10, and 5 percent. Figure 3 gives examples of the two lowest resolutions.

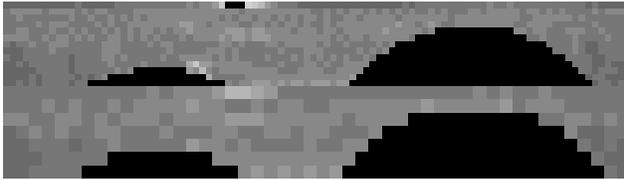

Figure 3: Two very low resolution normalized and masked irises made from the image of Figure 2. The upper one is 13x96 (10% of original resolution in both directions) and the lower one is 7x48 (5% of original resolution in both directions).

## 6. Results

Figures 4, 5, and 6 show imposter histograms for 100%, 30%, and 5% resolution images, respectively. The narrow black bars are an overlay of the binomial distribution having the same mean and standard deviation as the histogram.

The mean, standard deviation, and degrees of freedom for all resolutions are given in Table 1. The degrees of freedom are calculated from Equation (2) with p equal to the mean.

Figure 7 is a plot of degrees of freedom versus resolution.

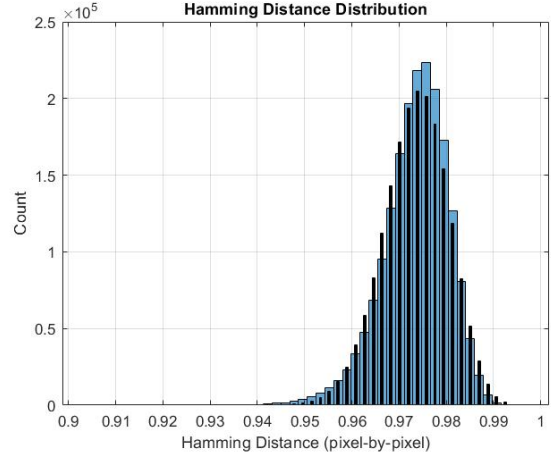

Figure 4: Imposter Histogram of 128x960 normalized and masked iris images with corresponding binomial distribution overlay (narrow black bars)

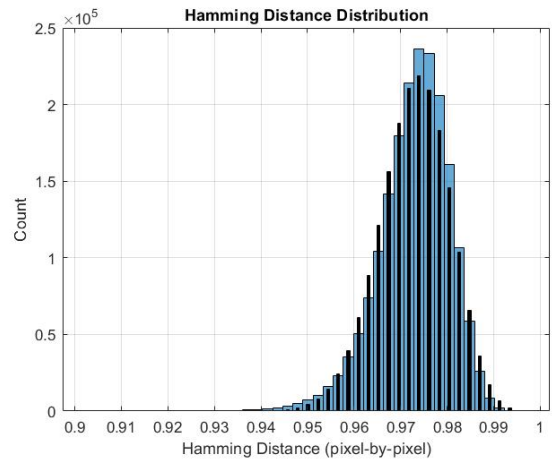

Figure 5: Imposter Histogram of 39x288 normalized iris images (30% of original resolution) with corresponding binomial distribution overlay (narrow black bars)

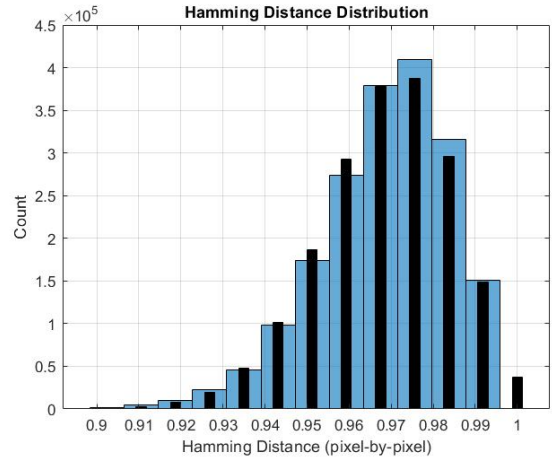

Figure 6: Imposter Histogram of 7x48 normalized iris images (5% of original resolution) with corresponding binomial distribution overlay (narrow black bars)

Table 1. Hamming Distance Comparison Results

| Resolution (percentage of 128x960) | Mean Hamming Distance | Std. Dev. Of Hamming Distance | Degrees of Freedom |
|---|---|---|---|
| 100 | 0.973508055 | 0.006933829 | 536 |
| 80 | 0.973432739 | 0.006990415 | 529 |
| 50 | 0.973239352 | 0.007175015 | 506 |
| 40 | 0.973124211 | 0.007317235 | 488 |
| 30 | 0.972883953 | 0.007562925 | 461 |
| 20 | 0.972312082 | 0.008135030 | 407 |
| 10 | 0.970584522 | 0.010361062 | 266 |
| 5 | 0.968513619 | 0.015774466 | 123 |

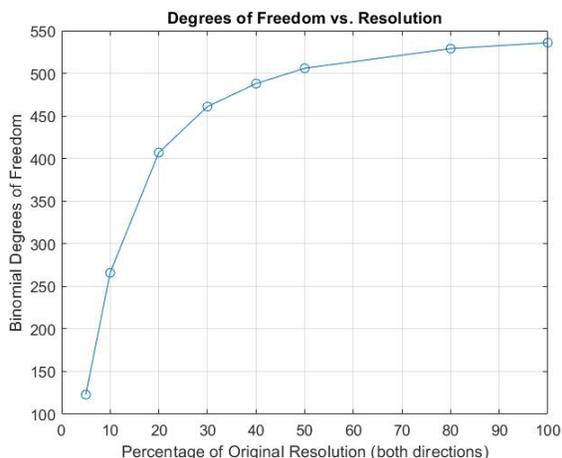

Figure 7: Binomial Degrees of Freedom versus Image Resolution

## 7. Discussion of Results

By doing pixel-by-pixel comparison of high-quality iris images, retaining both amplitude and phase information in the comparison, this investigation shows that there are at least 536 binomial degrees of freedom in the human iris. Only a single image of each eye is used. Uneven illumination, distortion caused by imperfect segmentation and normalization, imperfect eyelid/eyelash segmentation, and other "noise" that causes images of the same eye to vary significantly at different times were ignored. Thus, the pixel-by-pixel comparison method used in this investigation is impractical for iris recognition in the real world. But the results of this investigation do show that better methods of iris comparison with much higher degrees of freedom than current methods are theoretically possible.

## 8. Acknowledgements

The author thanks:

- Adam Czajka, Kevin Bowyer, et al at Notre Dame University for loan of the essential iris images [4], their whole body of excellent iris recognition work, and expert communication that helped the author to determine what research was worth doing.
- My colleagues at Eyelock LLC for consistently pushing the author and one another to make iris recognition better.